# Expert-Augmented Machine Learning

Efstathios D Gennatas[a]*, Jerome H Friedman[b], Lyle H Ungar[c], Romain Pirracchio[d], Eric Eaton[c], Lara G Reichmann[e], Yannet Interian[e], José Marcio Luna[f], Charles B Simone 2nd[g], Andrew Auerbach[h], Elier Delgado[i], Mark J van der Laan[j], Timothy D Solberg[a], Gilmer Valdes[a]

[a]University of California San Francisco, Department of Radiation Oncology
[b]Stanford University, Department of Statistics
[c]University of Pennsylvania, Department of Computer and Information Science
[d]University of California San Francisco, Department of Anesthesia and Perioperative Care
[e]University of San Francisco, Data Institute
[f]University of Pennsylvania, Department of Radiation Oncology
[g]New York Proton Center, Department of Radiation Oncology
[h]University of California San Francisco, Division of Hospital Medicine
[i]Innova Montreal Inc
[j]University of California Berkeley, Division of Biostatistics

*Current address: University of California, Department of Epidemiology and Biostatistics

Correspondence and requests for materials should be addressed to:
efstathios.gennatas@ucsf.edu

## Abstract

Machine Learning is proving invaluable across disciplines. However, its success is often limited by the quality and quantity of available data, while its adoption is limited by the level of trust afforded by given models. Human vs. machine performance is commonly compared empirically to decide whether a certain task should be performed by a computer or an expert. In reality, the optimal learning strategy may involve combining the complementary strengths of man and machine. Here we present Expert-Augmented Machine Learning (EAML), an automated method that guides the extraction of expert knowledge and its integration into machine-learned models. We used a large dataset of intensive care patient data to derive 126 decision rules that predict hospital mortality. Using an online platform, we asked fifteen clinicians to assess the relative risk of the subpopulation defined by each rule compared to the total sample. We compared the clinician-assessed risk to the empirical risk and found that while clinicians agreed with the data in most cases, there were notable exceptions were they over- or under-estimated the true risk. Studying the rules with greatest disagreement, we identified problems with the training data, including one miscoded variable and one hidden confounder. Filtering the rules based on the extent of disagreement between clinician-assessed risk and empirical risk, we improved performance on out-of-sample data and were able to train with less data. EAML provides a platform for automated creation of problem-specific priors which help build robust and dependable machine learning models in critical applications.

## Significance Statement

Machine Learning is increasingly used across fields to derive insights from data, which further our understanding of the world and help us anticipate the future. The performance of predictive modeling is dependent on the amount and quality of available data. In practice, we rely on human experts to perform certain tasks and on machine learning for others. However, the optimal learning strategy may involve combining the complementary strengths of man and machine. We present Expert-Augmented Machine Learning, an automated way to automatically extract problem-specific human expert knowledge and integrate it with machine learning to build robust, dependable and data-efficient predictive models.

# Introduction

Machine learning (ML) algorithms are proving increasingly successful in a wide range of applications but are often data-inefficient and may fail to generalize to new cases. In contrast, humans are able to learn with significantly less data by using prior knowledge. Creating a general methodology to extract and capitalize on human prior knowledge is fundamental for the future of ML. Expert systems, introduced in the 1960s and popularized in the 1980s and early 1990s, were an attempt to emulate human decision-making in order to address Artificial Intelligence problems (1). They involved hard-coding multiple if-then rules laboriously designed by domain experts. This approach proved problematic because a very large number of rules was usually required, and no procedure existed to generate them automatically. In practice, such methods commonly resulted in an incomplete set of rules and poor performance. The approach fell out of favor and attention has since been focused mainly on ML algorithms requiring little to no human intervention. More recently, the PROGnosis RESearch Strategy (PROGRESS) Partnership of the UK's Medical Research Council has published a series of recommendations to establish a framework for clinical predictive model development, which emphasize the important of human expert supervision of model training, validation and updating (2)(3).

Learning algorithms map a set of features to an outcome of interest by taking advantage of the correlation structure of the data. The success of this mapping will depend on several factors, other than the amount of actual information present in the covariates (aka features, aka independent variables), including the amount of noise in the data, the presence of hidden confounders and the number of available training examples. Lacking any general knowledge of the world, it is no surprise that current ML algorithms will often make mistakes that would appear trivial to a human. For example, in a classic study, an algorithm trained to estimate the probability of death from pneumonia labeled asthmatic patients as having a lower risk of death than non-asthmatics (4). While misleading, the prediction was based on a real correlation in the data: these patients were reliably treated faster and more aggressively, as they should, resulting in consistently better outcomes. Out of context, misapplication of such models could lead to catastrophic results (if, for example, an asthmatic patient was discharged prematurely or under-treated). In a random dataset collected to illustrate the widespread existence of confounders in medicine, it was found that colon cancer screening and abnormal breast findings were highly correlated to the risk of having a stroke, with no apparent clinical justification (5). Unfortunately, superior performance on a task as measured on test sets derived from the same empirical distribution, is often considered as evidence that real knowledge has been captured by a model. In a recent study, *CheXNet: Radiologist-Level*

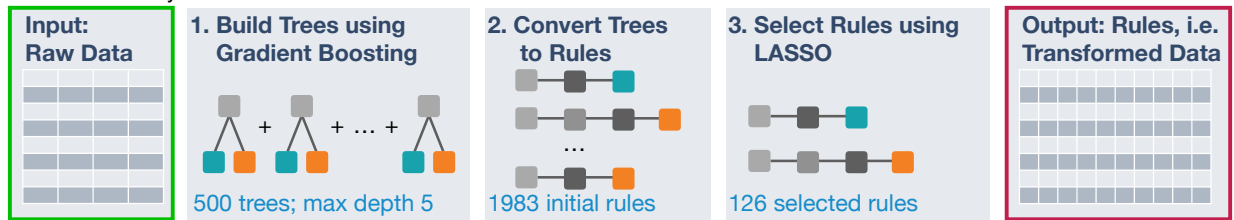
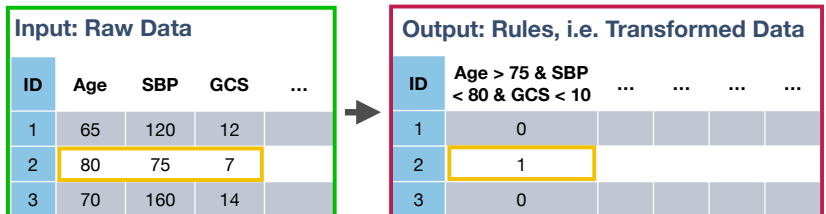

**Figure 1.** Overview of the methods. RuleFit involves 1) training a gradient boosting model on the input data, 2) converting boosted trees to rules by concatenating conditions from the root node to each leaf node and 3) training an L1-regularized (LASSO) logistic regression model. Each rule defines a subpopulation that satisfies all conditions in the rule. Clinician experts assess the mortality risk of the subpopulation defined by each rule compared to the whole sample on a web application. For each rule, delta ranking is calculated as the difference between the subpopulation's empirical risk as suggested by the data and the clinicians' estimate. A final model is trained by reducing the influence of those rules with highest delta ranking. This forms an efficient procedure where experts are asked to assess 126 simple rules of 3-5 variables each instead of assessing 24,508 cases with 17 variables each.

*Pneumonia Detection on Chest X-Rays with Deep Learning*, investigators observed that a Convolutional Neural Network (CNN) outperformed radiologists in overall accuracy (6). A subsequent study revealed that the CNN was basing some of its predictions on image artifacts that identified hospitals with higher prevalence of pneumonia or discriminated regular from portable radiographs (the latter is undertaken on sicker patients), while pathology present in the image was sometimes disregarded (7). It was also shown that performance declined when a model trained with data from one hospital was used to predict data from another (8).

Among the biggest challenges for ML in high-stakes applications like medicine, is to automatically extract and incorporate prior knowledge that allows ML algorithms to generalize to new cases and learn with less data. In this study, we hypothesized that combining the extensive prior knowledge of causal and correlational physiological relationships that human experts possess with a machine-learned model would increase model generalizability, i.e. out-of-sample performance. We introduce Expert-Augmented Machine Learning (EAML), a methodology to automatically acquire clinical priors for a given problem and incorporate them into ML model. The procedure allows training models with a) less data that are b) more robust to changes in the underlying variable distributions and c) resistant to performance decay with time. Rather than depending on hard-coded and incomplete rulesets, like the early expert systems did, or relying on potentially spurious correlations like current ML algorithms do, EAML guides the acquisition of prior knowledge to improve the final ML model. We demonstrate the value of EAML using the MIMIC dataset collected at the Beth Israel Deaconess Medical Center (BIDMC) between 2001 and 2012 and released by the PhysioNet team to predict mortality among intensive care unit (ICU) patients (9, 10).

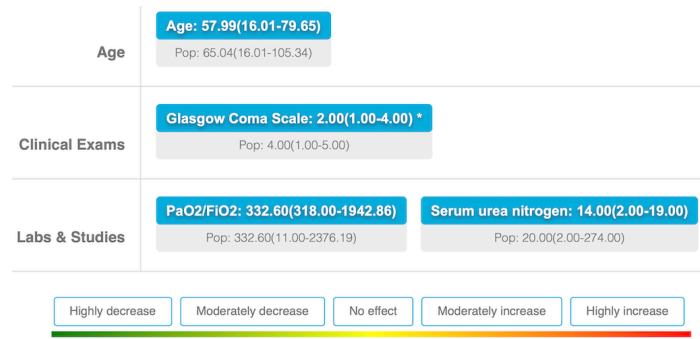

**Figure 2.** Example of a rule presented to clinicians. Age, Glasgow Coma Scale (1: <6, 2: 6-8, 3: 9-10, 4: 11-13, 5: 14-15), Ratio of Oxygen Blood concentration to Fractional Inspired Oxygen concentration (PaO2/FiO2) and Blood Urea Nitrogen concentration are the variables selected for this rule. The decision tree rules derived from gradient boosting, e.g. Age <= 73.65 & GCS <= 4, were converted to the form "median (range)", e.g. Age: 56.17 (16.01 – 73.65), for continuous variables and to the form "mode (included levels)" for categorical variables. Rules were presented in a randomized order, one at a time. The top line (blue box) displays the values for the subpopulation defined by the given rule. The bottom line (gray box) displays the values of the whole population. Participants were asked to assess the risk of belonging to the defined subpopulation compared to the whole sample using a 5-point system: highly decrease, moderately decrease, no effect, moderately increase, highly increase.

## EAML generates problem-specific priors from human domain experts

To automate the generation of problem-specific priors, we developed a multi-step approach (see Methods and summary in Figure 1). First, we trained RuleFit on the MIMIC-II ICU dataset collected at the BIDMC between 2001 – 2008 to predict hospital mortality using 17 demographic and physiologic input variables that are included in popular ICU scoring systems (11)(12)(13). This yielded 126 rules with nonzero coefficients. Using a 70% / 30% training / test split on the 24,508 cases, RuleFit achieved a test-set balanced accuracy of 74.4 compared to 67.3 for a Random Forest. Previously, Random Forest had been found to be the top performer among a library of algorithms on the MIMIC-II dataset (14). Subsequently, a committee of 15 clinicians at the University of California San Francisco were asked to categorize the risk of the subpopulations defined by each rule compared to the general population without being shown the empirical risk (Figure 2). On average, clinicians took 41 ± 19 minutes to answer 126 questions.

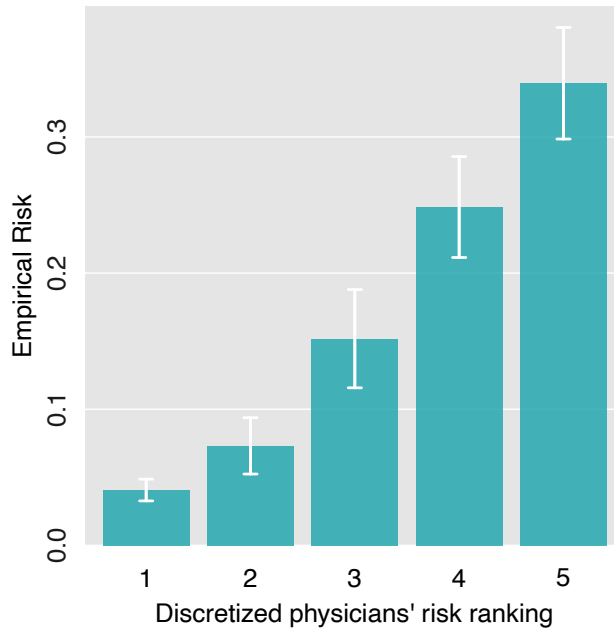

**Figure 3.** Mortality Ratio by average clinicians' risk ranking. Rules were binned into quintiles based on average clinicians' assessment. The mean empirical risk for each quintile was plotted. Error bars indicate 1.96 * standard error.

We calculated the average clinician assessment for each rule and ranked rules by increasing perceived risk, $Rank_p$. To check that we were successful in acquiring valid clinical information, we then binned the rules into five groups according to their ranking and plotted the empirical risk by group (Figure 3). There is a monotonic relationship between the average clinicians' ranking of a rule and its empirical risk (mortality ratio), as expected.

### Delta rank helps discover hidden confounders

The mortality ratio of patients within the subpopulation defined by each rule was used to calculate the empirical risk ranking of the rules, $Rank_e$. The delta ranking was defined as **ΔRank = $Rank_e$ – $Rank_c$** and is a measure of clinicians' disagreement with the empirical data.

| A. Clinician-estimated risk > Empirical risk | | ΔRank |
|---|---|---|
| 1 | Age=66.15 (16.5-89.3); PaO2/FiO2=332.6 (199.0-2304.8); HR=84.00 (0.0-106.0)); GCS=2 (1-4); Renal function=0 (0,1) | -49 |
| 2 | PaO2/FiO2=332.6 (224.0-955.0); GCS=5 (5); Age=80.9 (74.61-101.5); Renal function=0 (0) | -48 |
| 3 | GCS=2.0 (1.0-4.0); BUN=15.00 (2.0-24.0); Age=58.8 (16.8-75.2); PaO2/FiO2=332.6 (212.0 1942.9); HR=80.0 (0.00-92.0) | -47 |
| s4 | HR=80.00 (0.0-94.0); GCS=2 (1-4); BUN=15.0 (2.0-24.0); Age=62.7 (17.2-83.6)); PaO2/FiO2=332.6 (272.0-1942.9) | -47 |
| 5 | PaO2/FiO2=332.6 (318.6-2223.8); GCS=5 (3-5)); Age=81.2 (73.8-101.5); Renal function=0 (0) | -44 |
| 6 | HR=103.0 (93.0-171.0); GCS=1 (1-2); BUN=14.0 (2.0-23.0); PaO2/FiO2=345.0 (272.0-1939.3) | -43 |
| **B. Clinician-estimated risk < Empirical risk** | | **ΔRank** |
| 7 | GCS=5 (3-5); Bilirubin=2.7 (1.5-48.0); BUN=35.0 (20.0-248.0) | 37 |
| 8 | GCS=5 (4-5); BUN=44.0 (27.00-272.0); BP=91.0 (0.0-108.0) | 37 |
| 9 | PaO2/FiO2=496.5 (342.3-1942.9); HR=117.0 (107.0-171.0); BUN=13.0 (2.0-21.0) | 39 |
| 10 | PaO2/FiO2=122.9 (20.0-271.4); Age=53.8 (18.3-78.4); Bilirubin=3.6 (1.6-59.7) | 39 |
| 11 | GCS=5 (3-5); Bilirubin=4.0 (1.9-48.0); Renal function=1 (1,2,3,4) | 55 |
| 12 | Renal function=0 (0,1); PaO2/FiO2=470.0 (336.7-2304.8) | 56 |

**Table 1.** The top 5% rules in which the clinicians perceived risk is greater (A) and less (B) than the empirical risk. Rules have been color-coded to indicate similar concepts. Variables likely to have driven the response are highlighted in red. Values are shown as *Variable=median (range)*.

The distribution of ΔRank is shown in Figure S1. We hypothesized that those rules where ΔRank was outside the 90% confidence interval were likely to indicate either that clinicians misjudged the risk of the given subpopulations or that hidden confounders were modifying the risk. This hypothesis is based on the fact that the rules were created by the ML model based on empirical risk, while clinicians were estimating risk of each subpopulation based on medical knowledge and experience. We first analyzed those rules where the empirical ranking was significantly lower than the clinicians' perceived ranking (1A). For rules 1, 3, and 4 clinicians estimated that patients with a lower heart rate (HR) and Glasgow Coma Scale (GCS) score below 13 (in the original scale) are at higher risk than that supported by the data. For rules 2 and 5, clinicians appeared to overestimate the mortality risk of old age. Although it is true that older patients are generally at higher risk (11)(12)(15), the data suggests that being over 80 years old does not automatically increase one's risk of death in the ICU, if their physiology is not otherwise particularly compromised. Finally, the last rule in Table 1A indicates the discovery of a hidden confounder: intubation status. Intubated patients, whose responsiveness could not be assessed, were assigned the lowest possible GCS score in the MIMIC dataset. This was confirmed in correspondence with the PhysioNet team. Such a score would normally suggest a gravely ill patient who is unresponsive to external stimuli. Because the intubation status had not been initially collected, we reconstructed the same group of patients using MIMIC-III data and verified the miscoding (10). Patients with a GCS less than 8 who are not intubated (N = 1236) have a mortality risk of 0.28. Conversely, intubated patients (N = 6493) have a much lower mortality ratio of 0.19. The fact that intubated patients have been assigned the lowest possible GCS in the MIMIC-II dataset has largely been ignored in the literature. It was briefly mentioned by the PhysioNet team in the calculation of the SOFA score in the MIMIC-III dataset (16).

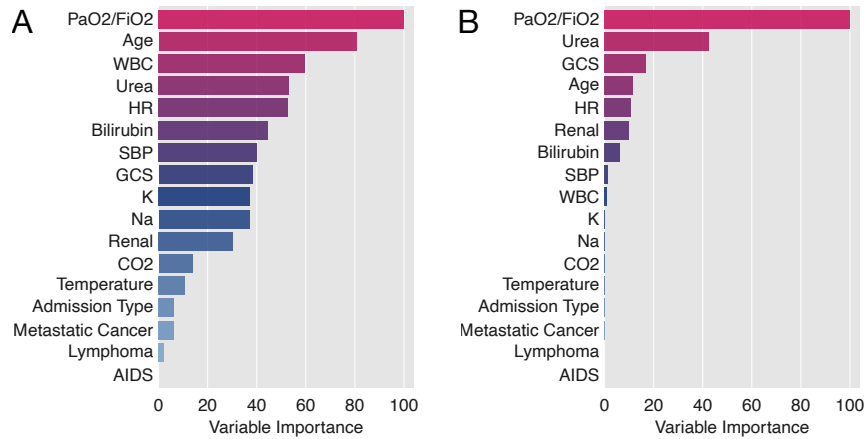

**Figure 4.** Variable importance estimated using a Random Forest model predicting mortality (A), and clinicians' assessments (B). While PaO2/FiO2 is the most important variable in both cases, in the former case it is used to learn intubation status, while in the latter clinicians are responding based on its physiological influence on mortality.

Table 1B shows the top 5% of the rules where the experts' ranking is lower than the empirical ranking. Here we find that clinicians have underestimated the influence of high blood urea nitrogen (BUN) or high bilirubin (rules 7, 8, 10, 11), although it is known that these variables affect mortality (17)(18)(19). The disagreement with the rules 9 and 12 allowed us to identify another important issue with the data: clinicians assigned a lower risk to patients with high ratio of arterial oxygen partial pressure to fractional inspired oxygen (PaO2/FiO2) than is supported by empirical data. In MIMIC-II, 54% of patients had missing values for PaO2/FiO2. After imputation with the mean, they were assigned a value of 332.60, which is very close to the value used by the rules in Table 1B (342.31 and 336.67 respectively). The notion that oxygen ratio missing values are not random and that imputation with the mean can cause models to predict incorrect risk near the mean oxygen ratio was pointed out to us by complementary work that predates ours by Rich Caruana and Sarah Tan [Personal Communication]. 94.2% of patients (N = 14430) with missing values for PaO2/FiO2 were not intubated, while 60.35% of patients with values for PaO2/FiO2 were intubated. Patients that were not intubated and had a PaO2/FiO2 greater than 336.67 had a mortality ratio of 0.046, which would agree with the clinicians' assessment. In contrast, patients that were intubated and had a PaO2/FiO2 greater than 336.67 had a mortality ratio of 0.13. Since this is approximately 60% of patients, they dominated the mortality risk on these rules (e.g. 0.10 for the last rule on Table 1B). As such, clinicians are again estimating risk based on their understanding of the effects of PaO2/FiO2 on mortality, while the algorithm has learned the effect of a hidden confounder; intubated vs. not intubated. To confirm this, we predicted intubation status in MIMICIII patients from the other covariates and achieved 97% mean accuracy using 10-fold cross-validation. This is especially troublesome because PaO2/FiO2 was selected by Random Forest as the most important variable in predicting mortality and was also selected as the most important variable driving clinicians' answers (Figure 4). The underlying reason in each case is however very different, as the algorithm is using PaO2/FiO2 as a proxy of intubation while clinicians are answering based on their understanding of physiology.

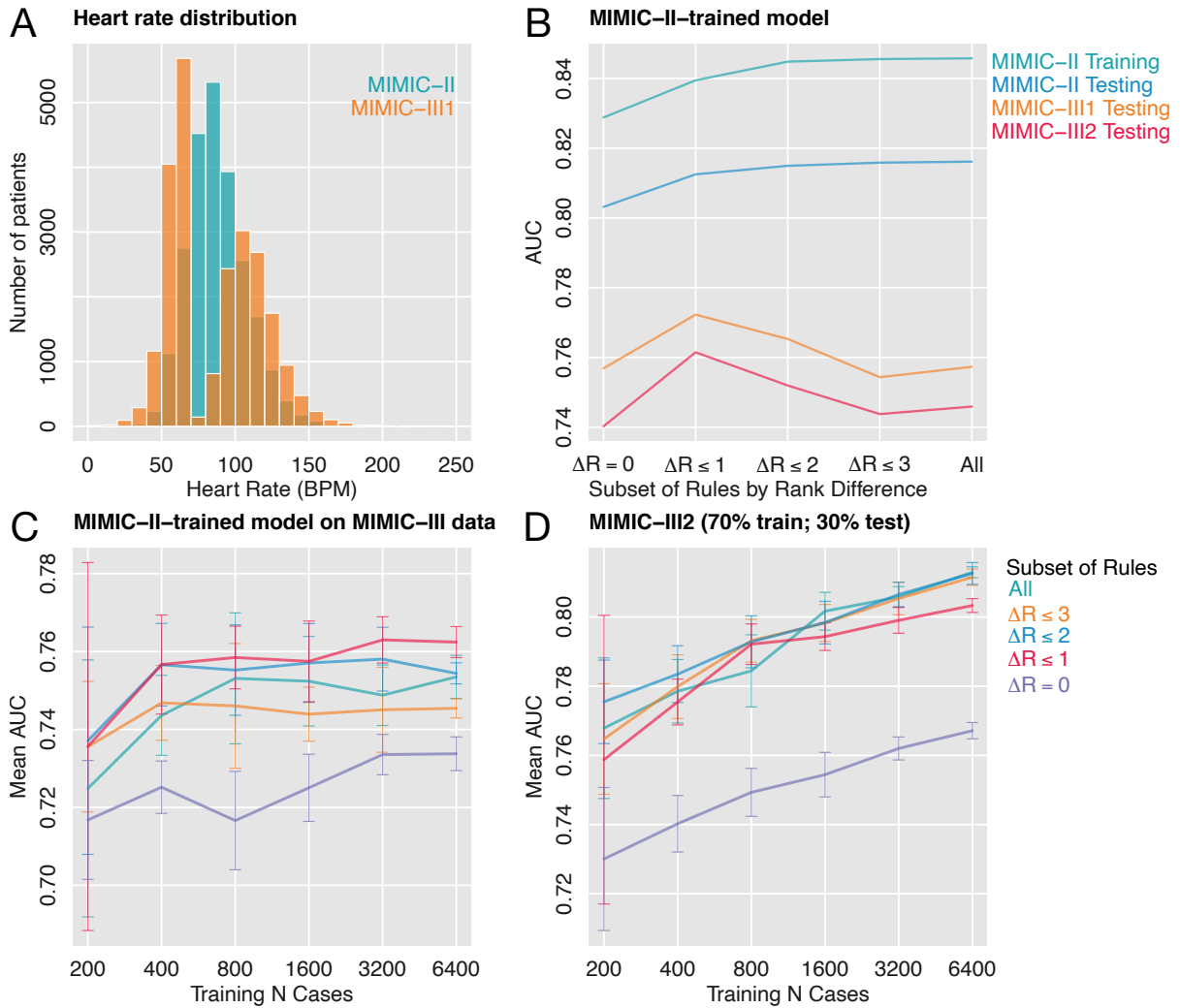

**Figure 5.** Example of variable shift: heart rate distribution of the same set of patients from MIMIC-II and MIMIC-III1 (A). Models were trained on MIMIC-II data using different subsets of rules defined by the extent of clinicians' agreement with the empirical risk (delta ranking cut in 5 bins, ΔR) (B). Mean AUC of models trained on MIMIC-II and tested on MIMIC-III (C) and models trained and tested on MIMIC-III (D). Subsamples of different sizes were used for each subset of rules defined by ΔR to test the hypothesis that eliminating bad rules helps the algorithm train with less data. Error bars represent 1 standard deviation across 10 stratified subsamples.

## Expert-Augmented Machine Learning improves out-of-sample performance

The MIMIC dataset was well suited for us to test whether MediForest + EAML can make models more robust to variable shifts or decay of accuracy with time. We built models combining clinicians' answers and the MIMIC-II dataset (collected from 2001 to 2008). We then evaluated these models on two sets of the MIMIC-III data: MIMIC-III1, which utilizes the same patients as in MIMIC-II but has different values of the input variables due to recoding of the underlying tables by the PhysioNet project, and MIMIC-III2 (collected from 2008 to 2012), which consists of new patients treated in the four years that followed the acquisition of MIMIC-II. Figure 5A illustrates an example of a variable distribution change from MIMIC-II to MIMIC-III1 (i.e. on the same cases).

Figure 5B illustrates the performance of models trained on 70% of MIMIC-II and evaluated on MIMIC-II (30% random subsample), MIMIC-III1 and MIMIC-III2. To demonstrate the effect of clinicians' knowledge, we first organized the rules into 5 categories according to a histogram of the absolute value of ΔRank, with ΔR = 0 reflecting those rules in which clinicians agreed the most with the empirical data and 5 the least. (In this text, we use ΔRank to refer to the difference between expert-assessed risk and empirical risk and ΔR to refer to the same measure after it has been cut into five bins). The effect of building different models by serially removing rules with increasing ΔR is illustrated in Figure 5B. This process can be considered as a *hard EAML*, where those rules that disagree more than a certain threshold are infinitely penalized (i.e. discarded) while those below the threshold are penalized by a constant. Since these rules were selected by RuleFit using the empirical distribution on MIMIC-II, getting rid of rules adversely affects performance (AUC) in the training data and in the testing set which originates from the same empirical distribution (Figure 5B). A different scenario emerges when these models are tested on both MIMIC-III1 and MIMIC-III2. In this case, penalizing those rules where clinicians disagree the most with the empirical data improves performance. When only rules with ΔR = 0 are left (N = 53 of 126 rules), however, performance decreases (Figure 5B). This suggests a tradeoff between using better rules to build the models (those in which clinician agree with the empirical risk) and oversimplifying the model (if only rules with ΔR = 0 are used). Therefore, better results might be obtained if we acquired clinicians' answers for all 2000 rules and not just the 126 selected by LASSO. The tradeoff here is time needed to collect experts' assessments.

Additionally, in Figure 5C we note that models with the highest accuracy can be obtained with half the data if clinicians' answers are used to limit rules used for training: models built with rules from groups 1 and 2, i.e. where ΔR ≤ 1, saturate around 400 patients while those built with all the rules need around 800 patients. Wilcoxon tests comparing performance of models trained on 6400 cases (saturation) using only rules with ranking difference ≤ 1 versus all rules show the reduced rule set results in significantly better AUC (W = 9, p-value = 0.00105) and balanced accuracy (W = 4, p-value = 0.00013). This effect is not present if the model is trained and tested on MIMIC-III data (Figure 5D). Figure 5 B-D exemplify the difficulties and limitations of selecting the best models using cross-validated errors estimated from the empirical distribution. Upon covariate shifts and data acquired at a different time (possibly reflecting new interventions and treatments, etc.), model selection using cross-validation from the empirical distribution is no longer optimal because spurious correlations found in the empirical distribution are likely to change. Since true causal knowledge does not change, our results suggest that this knowledge is being extracted from clinicians (e.g. evaluation of $PaO_2/FiO_2$ by clinicians). Finally, similar results can be obtained if instead of using the hard version of EAML, we use a soft version (SI Appendix).

## Discussion

Despite increasing success and growing popularity, ML algorithms can be data-inefficient and often generalize poorly to unseen cases. We have introduced EAML, the first methodology to automatically extract problem-specific clinical prior knowledge from experts and incorporate it into ML models. Related previous work had attempted to predict risk based on clinicians' assessment of individual cases using all available patient characteristics with limited success (20). Here, in contrast, we transformed the raw physiologic data into a set of simple rules and asked clinicians to assess the risk of subpopulations defined by those rules relative to the whole sample. We showed that utilizing this extracted prior knowledge allows: a) discovery of hidden confounders and limitations of clinicians' knowledge b) better generalization to changes in the underlying feature distribution c) improved accuracy in the face of time decay, d) training with less data and e) illustrating the limitations of models chosen using cross-validation estimated from the empirical distribution. We used the MIMIC dataset from the PhysioNet project (9)(10), a large dataset of intensive care patients, to predict hospital mortality. We showed that EAML allowed the discovery of a hidden confounder (intubation) that can change the interpretation of common variables used to model ICU mortality in multiple available clinical scoring systems – APACHE (11), SAPS II (21), or SOFA (13). Google Scholar lists over ten thousand citations of PhysioNet's MIMIC dataset as of December 2018, with approximately 1600 new papers published every year. Conclusions on treatment effect or variable importance using this dataset should be taken with caution, especially since intubation status can be implicitly learned from the data, as shown in this study, even though the variable was not recorded. Moreover, we identified areas where clinicians' knowledge may need evaluation and possibly further training, such as the case where clinicians overestimated the mortality risk of old age in the absence of other strong risk factors. Further investigation is warranted to establish whether clinicians' perceived risk is negatively impacting treatment decisions.

We have built EAML to incorporate clinicians' knowledge along with its uncertainty into the final ML model. EAML is not merely a different way of regularizing a machine-learned model but is designed to extract domain knowledge not necessarily present in the training data. We have shown that incorporating this prior knowledge helps the algorithm generalize better to changes in the underlying variable distributions, which, in this case, happened after a rebuilding of the database by the PhysioNet Project. We have also demonstrated that we can train models more robust to accuracy decay with time. Preferentially using those rules where clinicians agree with the empirical data not only produces models that generalize better, but it does so with considerably less data (N = 400 versus N = 800). This result can be of high value in multiple fields where data is scarce and/or expensive to collect. We also demonstrated the limitation of selecting models using cross-validated estimation from within the empirical distribution. We showed that there is no advantage in incorporating clinicians' knowledge if the test set is drawn from the same distribution as the training. However, when the same model was tested in a population whose variables had changed or that were acquired at a later time, including clinicians' answers improved performance and made the algorithm more data-efficient.

The MIMIC dataset offered a great opportunity to demonstrate the concept and potential of EAML. A major strength of the dataset is the large number of cases, while one of the main weaknesses is that all cases originated from a single hospital. We were able to show the benefit of EAML in the context of feature coding changes and time decay (MIMIC-III1 and

MIMIC-III2). However, proper application of EAML requires independent training, validation, and testing sets, ideally from different institutions. Crucially, an independent validation set is required in order to choose the best subset of rules (hard EAML) or the lambda hyperparameter (soft EAML). If the validation set has the same correlation structure between the covariates and outcome as the training set), cross-validation will choose a lambda of 0 provided there are enough data points. However, if the validation set is different from the training set, then incorporating expert knowledge will help and the tuning will result in lambda greater than 0. This is the same for any ML model training where hyperparameter tuning cannot be effectively performed by cross-validation of the training set if that set is not representative of the whole population of interest, which is most commonly the case in clinical datasets. One of the biggest contributions of this paper is showing the risk of using a validation set that has been randomly subsampled from the empirical distribution and as such contains the same correlations as the training data. Our team is preparing a multi-institutional EAML study to optimize the algorithm for real-world applications.

Finally, this work also has implications on the interpretability and quality assessment of ML algorithms. It is often considered that a tradeoff exists between interpretability and accuracy of ML models (22)(23). However, as shown by Friedman and Popescu (24), rule ensembles, and therefore EAML, are on average more accurate than Random Forest and slightly more accurate than Gradient Boosting in a variety of complex problems. EAML builds on RuleFit to address the accuracy-interpretability tradeoff in ML and allows one to examine all of the model's rule ahead of deployment, which is essential to building trust in predictive models.

## Methods

A more complete description of the study methods is available in the SI Appendix. Briefly, we used the publicly available MIMIC ICU dataset from the PhysioNet project to predict in-hospital mortality. The MIMIC dataset includes two releases: MIMIC-II, collected at BIDMC between 2001 – 2008 (9), and MIMIC-III (10), which includes the MIMIC-II cases after re-coding of some variables (which resulted in distribution shifts) plus new cases treated between 2008 – 2012. We split the data in four groups: 1. MIMIC-II training (70% of MIMIC-II stratified on outcome), 2. MIMIC-II testing (remaining 30% of MIMIC-II), 3. MIMIC-III1 (MIMIC-II cases after recoding), 4. MIMIC-III2 (new cases collected after 2008 not present in MIMIC-II). The seventeen input features consisted of demographics, clinical and physiological variables included in common ICU risk scoring systems.

The RuleFit procedure (24) was used to derive 126 decision rules made up of three to five input variables that predict mortality. These rules represent a transformation of the input variables to a Boolean matrix (i.e. True / False). For example, the rule "Age > 75 & systolic blood pressure < 80 & Glasgow Coma Scale < 10" will have a value of "1" for all patients that match each of these conditions and "0" otherwise, thus defining a subpopulation within the full sample. The RuleFit-derived rules were uploaded to a web application (www.mediforest.com). Fifteen hospitalists and ICU clinicians were asked to assess the relative mortality risk of patients belonging to the subgroup defined by each rule relative to the whole population by selecting one of five possible responses: highly decrease, "1", moderately decrease, "2", no effect, "3", moderately increase, "4", and highly increase, "5". Rules were ranked based on the empirical risk of their respective subpopulations ($Rank_e$) and by the mean clinician-assessed risk ($Rank_p$). The difference $\Delta Rank = Rank_p - Rank_e$ was calculated to represent the extent of agreement between the empirical data and the expert assessments and was used a) to identify problems in the training data and b) to regularize the final EAML model by penalizing rules with higher disagreement. All analysis and visualization were performed using the **rtemis** machine learning library (25).

## Data Availability

The software library used in this study is available on GitHub at https://github.com/egenn/rtemis. The code used to perform this study along with the rankings obtained from clinicians is available at https://github.com/egenn/EAML_MIMIC_ICUmortality.

## Acknowledgments

We thank Rich Caruana for stimulating discussions and two anonymous reviewers for their constructive feedback.

## Conflicts of interest

Authors report no conflicts of interest

## Author Contributions

**EDG**: Conceptualization, algorithm design, software, data analysis, manuscript
**JHF**: Conceptualization, algorithm design, manuscript

**LHU**: Conceptualization, manuscript

**RP**: Conceptualization, data analysis, manuscript

**EE**: Conceptualization, manuscript

**LGR**: Data analysis, manuscript

**YI**: Conceptualization, data analysis, manuscript

**CBS**: Conceptualization, data analysis, manuscript

**AA**: Conceptualization, data analysis, manuscript

**ED**: Web application, data analysis, manuscript

**MJvdL**: Conceptualization, manuscript

**TDS**: Conceptualization, manuscript

**GV**: Conceptualization, algorithm design, data analysis, project administration, manuscript

# Supplementary Information

## Methods

### Dataset

The publicly available MIMIC dataset from the PhysioNet project was used in this study. The project's Institutional Review Board (IRB) was approved by the Beth Israel Deaconess Medical Center (Boston, MA) and the Massachusetts Institute of Technology (Cambridge, MA). Patient consent was not sought because the study did not impact clinical care and protected health information was de-identified in compliance with the Health Insurance Portability and Accountability Act (HIPAA). Patient data collection was first performed from 2001-2008 (MIMIC-II) (1) while the Beth Israel Deaconess Medical Center (BIDMC) used the CareVue management software (Philips, Andover, MA). In 2013, the MIMIC-III dataset was released (2) that included a) the same patients in MIMIC-II but with many data elements regenerated from the raw data in a more robust manner and b) extra patients treated from 2008-2012. It is important to note that the BIDMC switched management software from CareVue to Metavision (iMDSoft, Wakefield, MA) in 2008. In this study we included ICU patients older than 15 years with a single admission per hospital stay, resulting in 24,508 cases from MIMIC-II and 44,010 for MIMIC-III. The target outcome was in-hospital mortality. The features used to build our prediction algorithms included 13 physiological variables (age, Glasgow Coma Scale, systolic blood pressure, heart rate, body temperature, PaO2/FiO2 ratio, urinary output, serum urea nitrogen level, white blood cell count, serum bicarbonate, sodium, potassium and bilirubin levels), type of admission (scheduled surgical, unscheduled surgical, or medical), and three underlying disease variables (acquired immunodeficiency syndrome, metastatic cancer, and hematologic malignancy derived from ICD-9 discharge codes). These variables were set to the worst value recorded in the first 24 hours as defined by Le Gall et al (3). They were selected because they are easy to acquire and are used in the majority of clinically available scores – APACHE (4), SAPS (3) , SOFA (5). After reviewing clinicians' answers, we extracted additional features from MIMIC-III, like intubation status, that were hypothesized to explain the disagreement between clinicians and the empirical data.

# EAML: Rule-Generation Algorithm

In order to extract information from clinicians, a ML model was constructed by applying the RuleFit algorithm (6) to the MIMIC-II dataset. RuleFit first uses Gradient Boosting (7) with hyperparameters selected to introduce diversity to obtain a large number of decision trees and converts them to a set of binary decision rules. For each case, each rule is either 1 or 0, if the patient satisfies the criteria. These rules are then used as input features in a LASSO model, which performs variable selection, effectively selecting the most important rules. Equation 1 shows the squared error loss function for the regression case for simplicity (in this study, all experiments used the negative binomial logistic loss function).

$$\hat{c}_0, \{\hat{c}_k\}_1^K = argmin_{\hat{c}_0, \{\hat{c}_k\}_1^K} \sum_{i=1}^{N} (y_i - c_0 - \sum_{k=1}^{K} c_k r_{ik})^2 + \lambda \sum_{k=1}^{K} |C_k| \qquad (1)$$

where $\hat{c}_0, \{\hat{c}_k\}_1^K$ are the coefficients of the rules, $r_{ik}$, for each observation i that we would like to obtain, $y_i$ is the outcome and $\lambda$ is the lasso shrinkage parameter. The indicator $r_{ik} = 1$ if observation i belongs to rule k or 0 otherwise. Future observations are predicted using Equation 2.

$$y = c_0 + \sum_{k=1}^{K} c_k r_k \qquad (2)$$

Note that since equation 1 solves the lasso problem, most coefficients $\{\hat{c}_k\}_1^K$ will be set to 0; approximately only 10% of coefficients will be nonzero (6). Friedman and Popescu demonstrated that even when modeling complicated functions, approximately 200 rules built from trees of depth 3 suffice to give models that are competitive in accuracy compared to state-of-the-art algorithms like Random Forests (8) and Gradient Boosting (7). In the present article we applied RuleFit to the MIMIC-II database to generate simple rules that were used as building blocks to extract knowledge from clinicians.

# Web Application: MediForest

A web application, called MediForest, was created to collect clinician knowledge. Hospitalists and ICU clinicians were contacted by email and compensated 100 USD for participation. Fifteen clinicians were invited, and all answered the 126 questions. Before participating, clinicians were asked to watch a video explaining the MediForest interface (https://youtu.be/

pqDnElOLoxw). In all cases, clinicians were tasked to assess the risk of a subgroup of patients defined by a given rule relative to the whole population by selecting one of five possible responses: highly decrease, "1", moderately decrease, "2", no effect, "3", moderately increase, "4", and highly increase, "5". The clinician-assessed risk, $R_k$, was defined as the average clinician response for each rule k. The standard deviation over clinicians' answers for each rule, $STDV_k$, was taken as a measurement of inter-clinician agreement. Rules were then ranked according to perceived risk from lowest to highest, $Rank_c$. This ranking represents the acquired clinician knowledge. Rules were also ranked according to the actual mortality ratio of patients within the rule, $Rank_e$. The difference between empirical and clinician-estimated risk was calculated, $\Delta R = Rank_e - Rank_c$. Figure S1 shows the distribution of $\Delta R$. Rules with $\Delta R$ outside the 90% confidence interval were investigated.

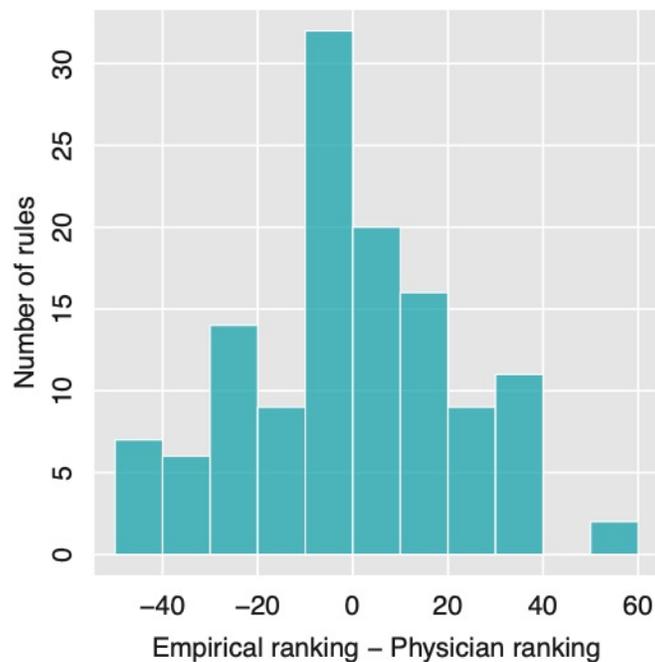

**Fig S1.** Distribution of the disagreement between clinician-derived and empirical risk ranking, $\Delta \text{Rank}$.

**Expert-Augmented Machine Learning**

In order to incorporate clinician knowledge (represented by ΔR and STDV) into the ML model, we designed the Expert Augmented Machine Learning (EAML) algorithm. In its most general form, EAML is a group-penalized regression (9) where each rule is penalized as a function of the clinicians' disagreement with the data, $\Delta R_k$, and a measurement of trust in this disagreement, $STDV_k$, such that:

$$\hat{c}_0, \{\hat{c}_k\}_1^K = \underset{\hat{c}_0, \{\hat{c}_k\}_1^K}{argmin} \sum_{i=1}^{N} L(y_i, c_0 + \sum_{k=1}^{K} c_k r_{ik}) + \lambda \sum_{k=1}^{K} f(\Delta R_k, STDV_k) \, |C_k|_m \quad (3)$$

where **L** is a general loss function and $|c|_m$ is the norm m of the vector **c**.

For instance, if we take $f(\Delta R_k, STDV_k) = \dfrac{|\Delta R_k|}{STDV_k}$ and m = 1, we have that those rules where clinicians disagree with the data, (higher $\Delta R_k$), get penalized more for bigger lambda. Additionally, those rules where clinicians disagree among themselves, characterized by a high $STDV_k$, get penalized less for the same $|\Delta R_k|$. Finally, the hyperparameter lambda controls the level of trust we place on clinician knowledge vs. the data. For λ >> 0, only rules with $|\Delta R_k|$ = 0 will be included in the model (complete trust on the clinicians' answer). For λ = 0, the clinicians' knowledge is discarded. In general, λ is a hyperparameter that should be defined as a function of the quality of the data and the clinicians' knowledge about a certain problem,

$$\lambda = f(Data, Expert\ knowledge)$$

and selected based on an independent validation set.

**Soft EAML**

Besides the hard version of EAML described in equation 3 of the main text, we also define a soft EAML:

$$\hat{c}_0, \{\hat{c}_k\}_1^K = \underset{c_0, \{c_k\}_1^K}{argmin} \sum_{i=1}^{N} L(y_i, c_0 + \sum_{k=1}^{K} c_k r_{ik}) + \lambda \sum_{k=1}^{K} \left(1 + \gamma \frac{|\Delta R_k|}{STDV_k + 4\max(STDV_k)_1^K}\right) |C_k|_2 \quad (4)$$

Equation 4 introduces the parameter γ to control the extra penalization introduced on top of ridge penalization that is applied to rules with greater ΔR. If γ = 0, then Ridge regression is recovered. If λ = 0, then regular linear regression is recovered regardless of γ. Additionally, the term $4\max(STDV_k)_1^K$, which indicates the maximum value of the standard deviation among all the rules, has been added to the denominator so that the maximum variation that the standard deviation can introduce is 20% of the original $|\Delta R_k|$. Figure S1 shows the effect of varying λ and γ on the training and testing sets MIMIC-II, MIMIC-III1, and MIMIC-III2. Figure S2 shows that in training data and testing data derived from the same distribution (MIMIC-II), no regularization (λ = 0) gives the best results but testing performance on MIMIC-III1 and MIMIC-III2 improves with γ > 0, which indicates the value of incorporating clinicians' answers.

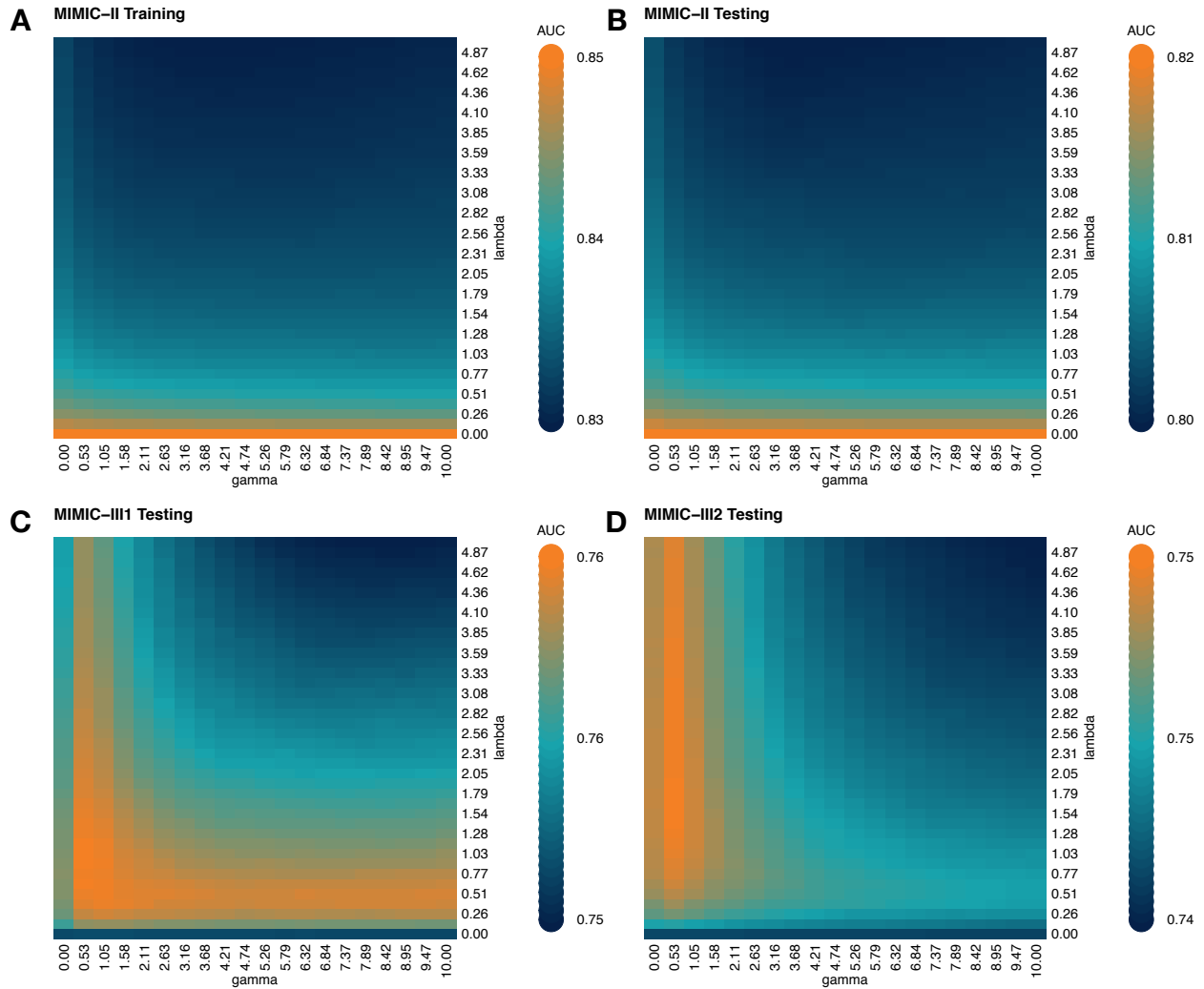

**Fig S2.** Soft EAML: Models were trained on 70% of the MIMIC-II dataset using varying combinations of hyperparameters γ (gamma) and λ (lambda). Model performance is shown on the training set (A), the 30% MIMIC-II left-out test set (B), MIMIC-III1 (C) and MIMIC-III2(D).

A summary of the methods is shown in Figure 1 of the main text. All analysis and visualization were performed using the **rtemis** machine learning library (10).